\documentclass{article}

\usepackage[final,nonatbib]{itml2019}

\usepackage[square,numbers]{natbib}
\usepackage[utf8]{inputenc} % allow utf-8 input
\usepackage[T1]{fontenc}    % use 8-bit T1 fonts
\usepackage{hyperref}       % hyperlinks
\usepackage{url}            % simple URL typesetting
\usepackage{booktabs}       % professional-quality tables
\usepackage{amsfonts}       % blackboard math symbols
\usepackage{nicefrac}       % compact symbols for 1/2, etc.
\usepackage{microtype}      % microtypography
\usepackage{amsthm}
\usepackage{amsmath}
\usepackage{xfrac}
\usepackage{wrapfig}
\usepackage{bbm}
\usepackage{subfig}
\usepackage{mathtools}

\usepackage{tikz}
\usepackage{verbatim}
\usepackage{ulem} %cross out text
\usetikzlibrary{decorations.pathreplacing}

\usepackage{xcolor}

\title{Information-Theoretic Perspective of Federated Learning}

\author{%
  Linara Adilova \\
  Fraunhofer Center for Machine Learning \\
  Fraunhofer IAIS, Germany \\
    \texttt{linara.adilova@iais.fraunhofer.de}
    \And
    Julia Rosenzweig \\
    Fraunhofer IAIS, Germany \\
    \texttt{julia.rosenzweig@iais.fraunhofer.de}
   \And
    Michael Kamp\\
    Monash University, Australia \\
    \texttt{michael.kamp@monash.edu}
}

\begin{document}

\maketitle

\begin{abstract}
An approach to distributed machine learning is to train models on local datasets and aggregate these models into a single, stronger model. A popular instance of this form of parallelization is federated learning, where the nodes periodically send their local models to a coordinator that aggregates them and redistributes the aggregation back to continue training with it. The most frequently used form of aggregation is averaging the model parameters, e.g., the weights of a neural network. However, due to the non-convexity of the loss surface of neural networks, averaging can lead to detrimental effects and it remains an open question under which conditions averaging is beneficial. In this paper, we study this problem from the perspective of information theory: We measure the mutual information between representation and inputs as well as representation and labels in local models and compare it to the respective information contained in the representation of the averaged model. Our empirical results confirm previous observations about the practical usefulness of averaging for neural networks, even if local dataset distributions vary strongly. Furthermore, we obtain more insights about the impact of the aggregation frequency on the information flow and thus on the success of distributed learning. These insights will be helpful both in improving the current synchronization process and in further understanding the effects of model aggregation.
\end{abstract}

\section{Introduction}
In distributed machine learning with decentralized data and communication constraints, federated learning has become a popular approach, particularly for training neural networks~\citep{mcmahan2017communication, kamp2018efficient}. The idea is to train models locally and periodically average their parameters --- assuming that all local networks have the same architecture. Local nodes then continue training with the average. This approach is well-understood in the convex case~\citep{kamp2014communication}, but can be arbitrarily bad in the non-convex case of training neural networks. So far, the approach has been shown to work in practice in several scenarios, given that the local models are initialized similarly. The existing research is approaching this problem via studying the geometry of loss surfaces \citep{keskar2016large, nguyen2017loss}, but it is yet hard to apply it directly to the distributed case.
Information theory is a statistical basement of data science, but the fact that many expressions we encounter in this context are analytically intractable is a limiting factor for its widespread application. \citet{tishby2015deep} were the first to apply information theory to deep learning. A neural network is seen as a Markov chain and the information about input data propagated through layers is decreasing with the deepness of the layer. The goal of the training is maximizing the information about the label contained in the representations in the network and compressing the information about the input as much as possible to still obtain sufficient statistics. Many others followed on this line of research, including \citet{tishby2017} and \citet{Saxe2018}.

We apply information theory to understand the aggregation process of deep neural networks in a federated learning setup. For that, we analyse the development of mutual information in the local  and global models with averaging as aggregation method. Mutual information is a quantity that measures how much can be learned about one random variable from another. It is defined in terms of Kullback-Leibler divergence, i.e., $MI(X,Y) \coloneqq KL(p(x,y) || p(x)p(y))$ where $p(x)$ and $p(y)$ are marginal probability distributions of $X$ and $Y,$ respectively, and $p(x,y)$ is the joint probability distribution. 
%So it reflects how much the probability distributions of $X$ and $Y$ are close to being independent. 
%In practice we use EDGE \citep{noshad2019scalable} to estimate $MI(X,Z)$ between the input $X$ and the representation $Z$ at the last hidden layer of the network as well as the mutual information $MI(Z,Y)$ between the representation and the label $Y$ for the local and averaged models and each of the local datasets. By estimating these two quantities in different scenarios of federated learning, e.g., for iid and non-iid local datasets, we shed light on the inner workings of averaging as aggregation method for federated learning. 
There exist known conditions for the beneficial application of averaging, e.g., frequent aggregation into a global model and iid distribution of the local datasets \citep{mcmahan2017communication}. We analyse the effect of the fulfillment of such conditions on the results of training. 
%In practice we analyze conditions of the beneficial application of averaging such as frequent aggregation into a global model and iid distribution of the local datasets \citep{mcmahan2017communication}. 
To the best of our knowledge, there are no other attempts yet to study federated learning from an information-theoretic perspective.

\section{Experiments}
For our experiments we choose the image classification task CIFAR10~\citep{krizhevsky2014cifar} and apply the convolutional network LeNet~\citep{lecun1998gradient} to learn it. For the distributed setup we take the case with two local models for more tractable analysis. Mutual Information is estimated at each aggregation with the EDGE technique~\citep{noshad2019scalable} for the local networks as well as for the global one. In the first setup we evaluate the impact of averaging once at the end of the training process. We analyze this with respect to the length of the training process and find that, the longer the training process, the less successful averaging is. The second setup is periodic averaging with redistributing the averaged model. We perform our experiments for the case of iid and non-iid local datasets, because according to the literature iid data results in better model quality \citep{mcmahan2017communication} and with non-iid datasets we can more clearly see how the information that is existent only in one of the local nodes propagates with the help of averaging.

The aggregation is performed every $100$ local batches. Averaging on the networks is performed weight-vise. The training setup chosen is mini-batch stochastic gradient descent with batch size $32$. Every local network is trained for $20$ epochs on its local dataset (of size $25000$).
We measure mutual information of the representation $Z$ with input $X$ and of the representation $Z$ with label $Y$. We consider as representation $Z$ the output of the last hidden fully connected layer of the network. We estimate information separately for each of the local datasets in all the cases.
The non-iid case was emulated via separating $10$ of the CIFAR10 classes into non intersecting groups of $5$ classes each and presenting each local node examples only from one of these groups.

\begin{figure}[ht]
\minipage{0.5\textwidth}
  \includegraphics[width=\linewidth]{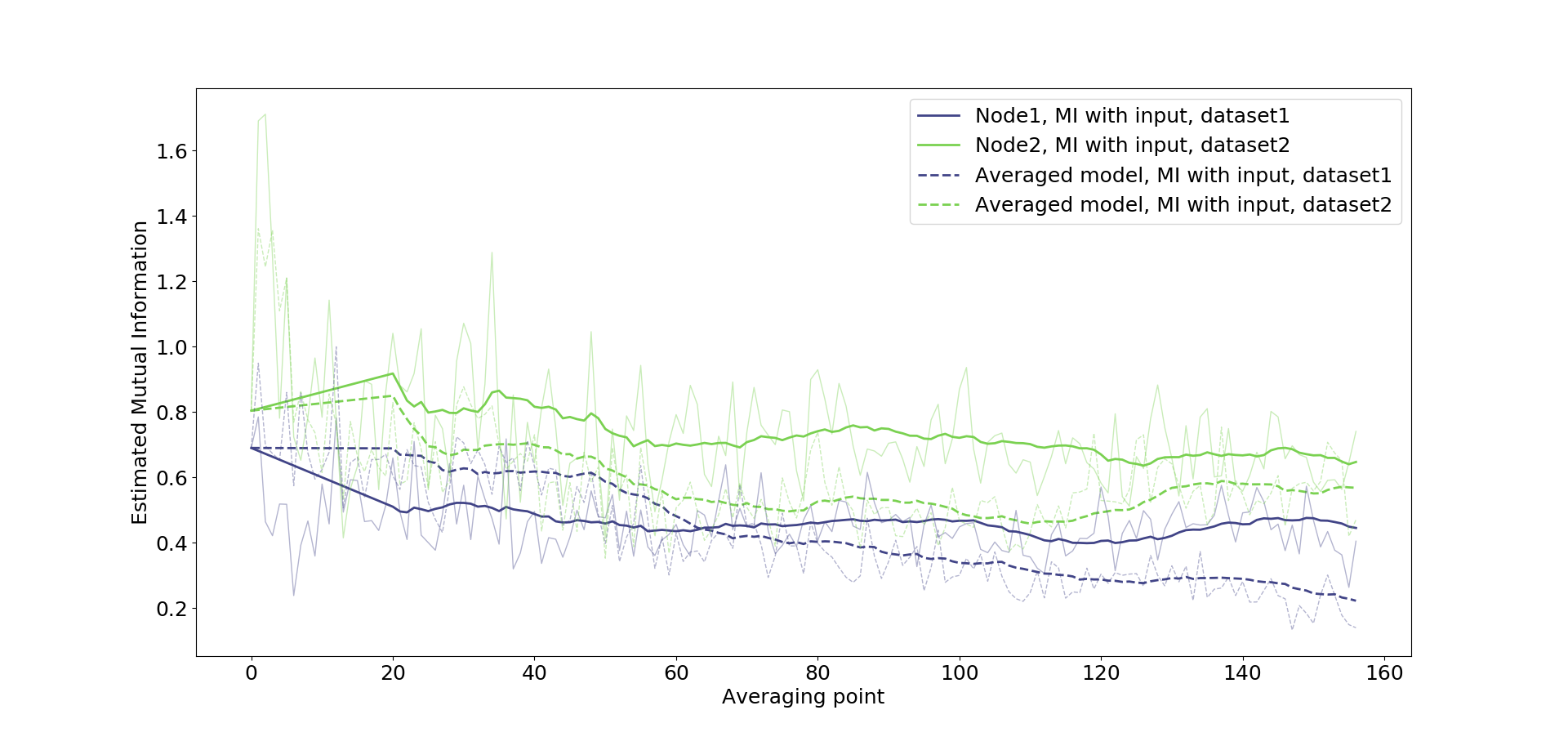}
\endminipage\hfill
\minipage{0.5\textwidth}
  \includegraphics[width=\linewidth]{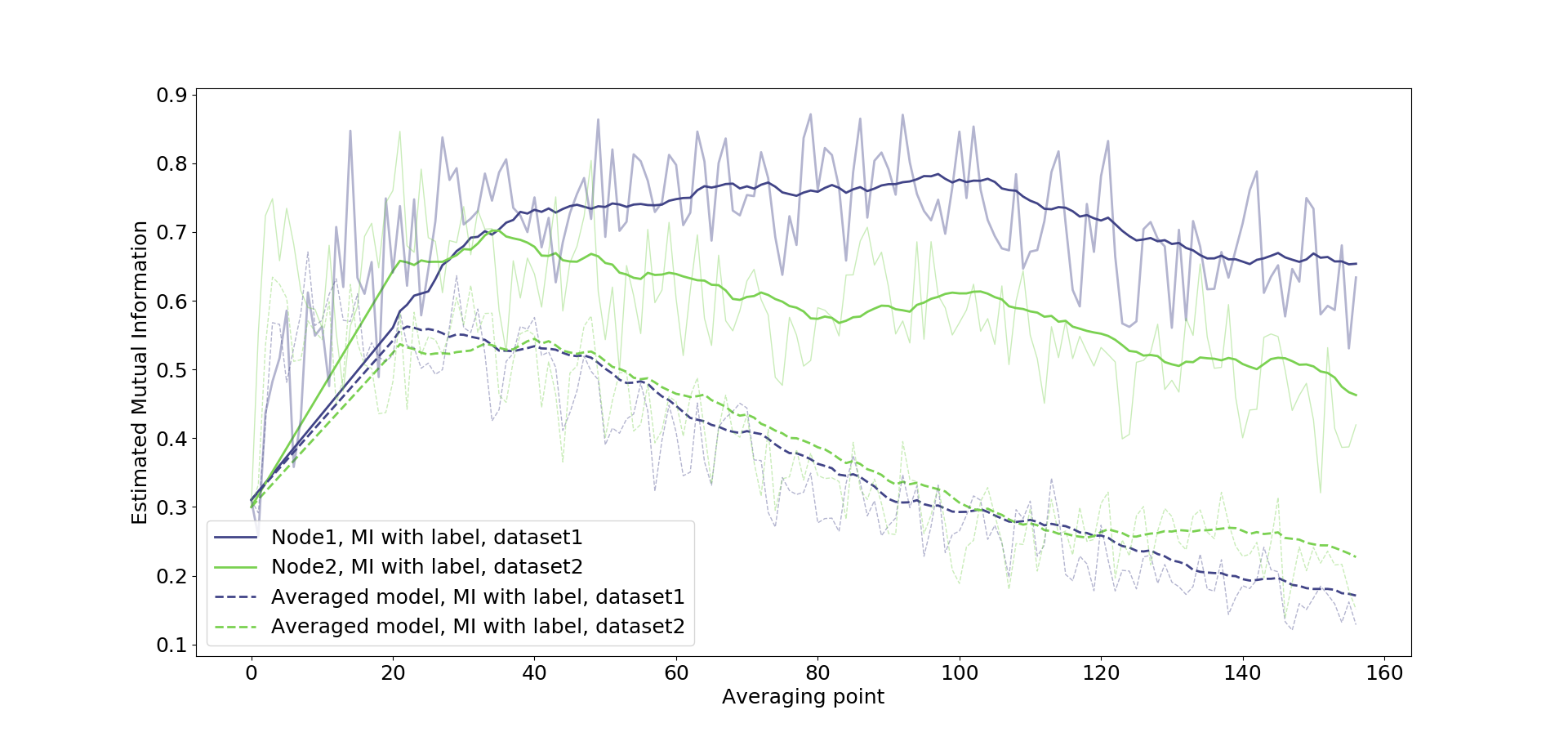}
\endminipage
\caption{Mutual information measured for the iid case without sending back the aggregated model. The estimated amount of information is shown for input $X$ (left) and label $Y$ (right) for each aggregation point.}
\label{fig:iid_nosync}
\end{figure}

\begin{figure}[ht]
\minipage{0.5\textwidth}
  \includegraphics[width=\linewidth]{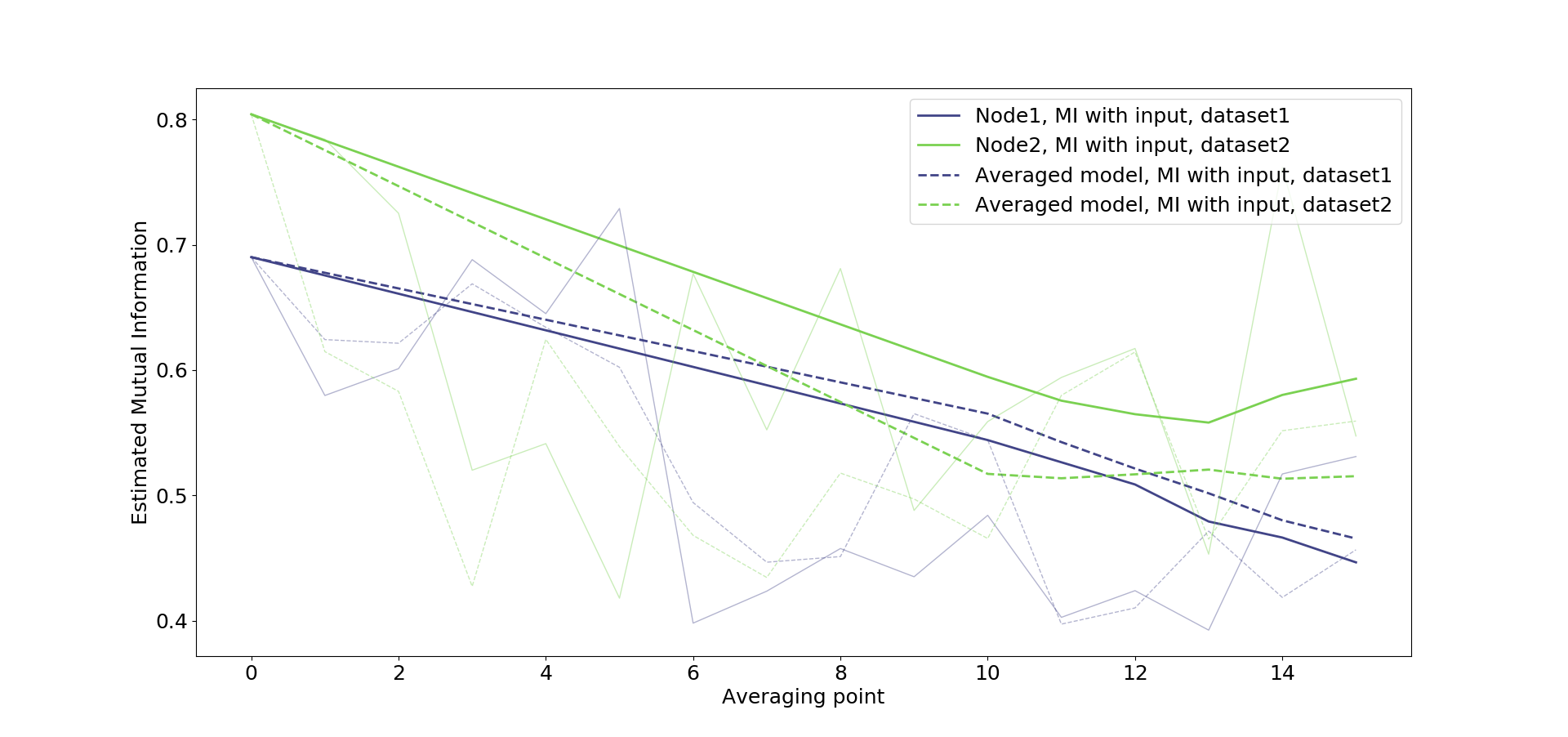}
\endminipage\hfill
\minipage{0.5\textwidth}
  \includegraphics[width=\linewidth]{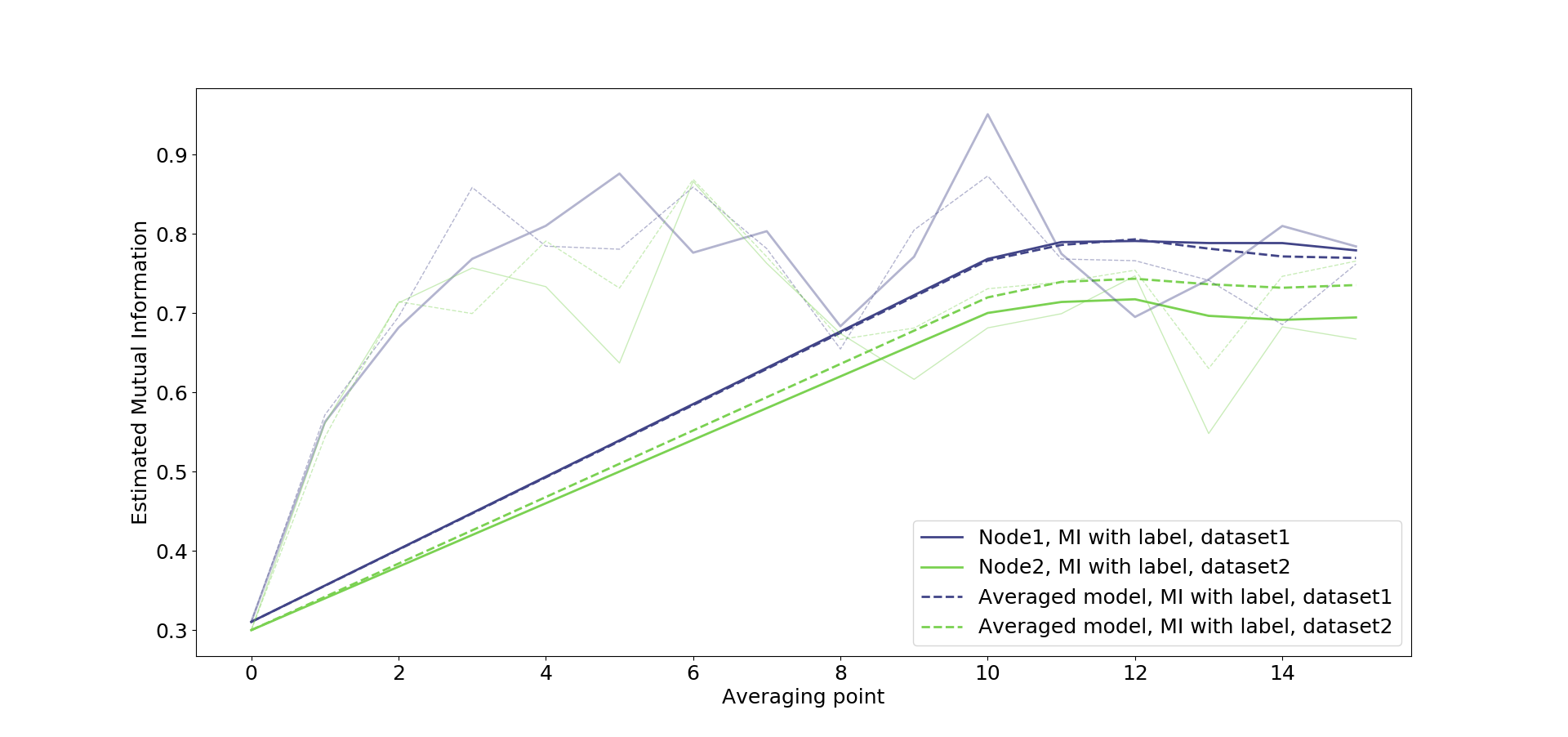}
\endminipage
\caption{Mutual information measured  for the iid case with aggregation process and continuing learning from the averaged point. The estimated amount of information is shown for input $X$ (left) and label $Y$ (right) for each aggregation point.}
\label{fig:iid_sync}
\end{figure}

In Figures~\ref{fig:iid_nosync} and \ref{fig:iid_sync} mutual information was measured in the setup with iid datasets. 
%Since each of the models gets an equivalent amount of examples for all the classes their fitting process should be approximately equivalent. Nevertheless, we see that the more information is learned by the representation about the labels, the less information is saved in the averaged model. 
After approximately $50$ aggregations without back-propagating the aggregated model, the information in the averaged model falls drastically. While each of the nodes eventually achieves more than $75\%$ training accuracy on the corresponding local dataset, the final averaged model has only $36\%$ of accuracy on each of the local datasets.
Since Figure~\ref{fig:iid_nosync} indicates that local models fail to be successfully averaged after the $10$-th aggregation, we perform an experiment with periodic synchronization having the period of synchronization set exactly to $10*100=1000$ local batches for training. The resulting measurements in Figure~\ref{fig:iid_sync} show that in this case mutual information with input behaves in approximately the same way, while mutual information with label does not drop anymore and even grows significantly. This also results in higher training accuracy of the averaged model -- more than $73\%$ on each of the local datasets in the end of the training. 
%At the same time sending the average back to the local nodes leads to slightly lower training accuracy of the local nodes themselves since their task is becoming harder with the training process being regularly interrupted. 
Interesting to note, that such a big period of synchronization is usually not used in the federated setup, nevertheless the experiments show that it can be beneficial. Thus, an analysis of information flow in the averaged model can help to identify a period for synchronization that is large enough and thus saving communication costs, but at the same moment allows for successful distributed training.

\begin{figure}[ht]
\minipage{0.5\textwidth}
  \includegraphics[width=\linewidth]{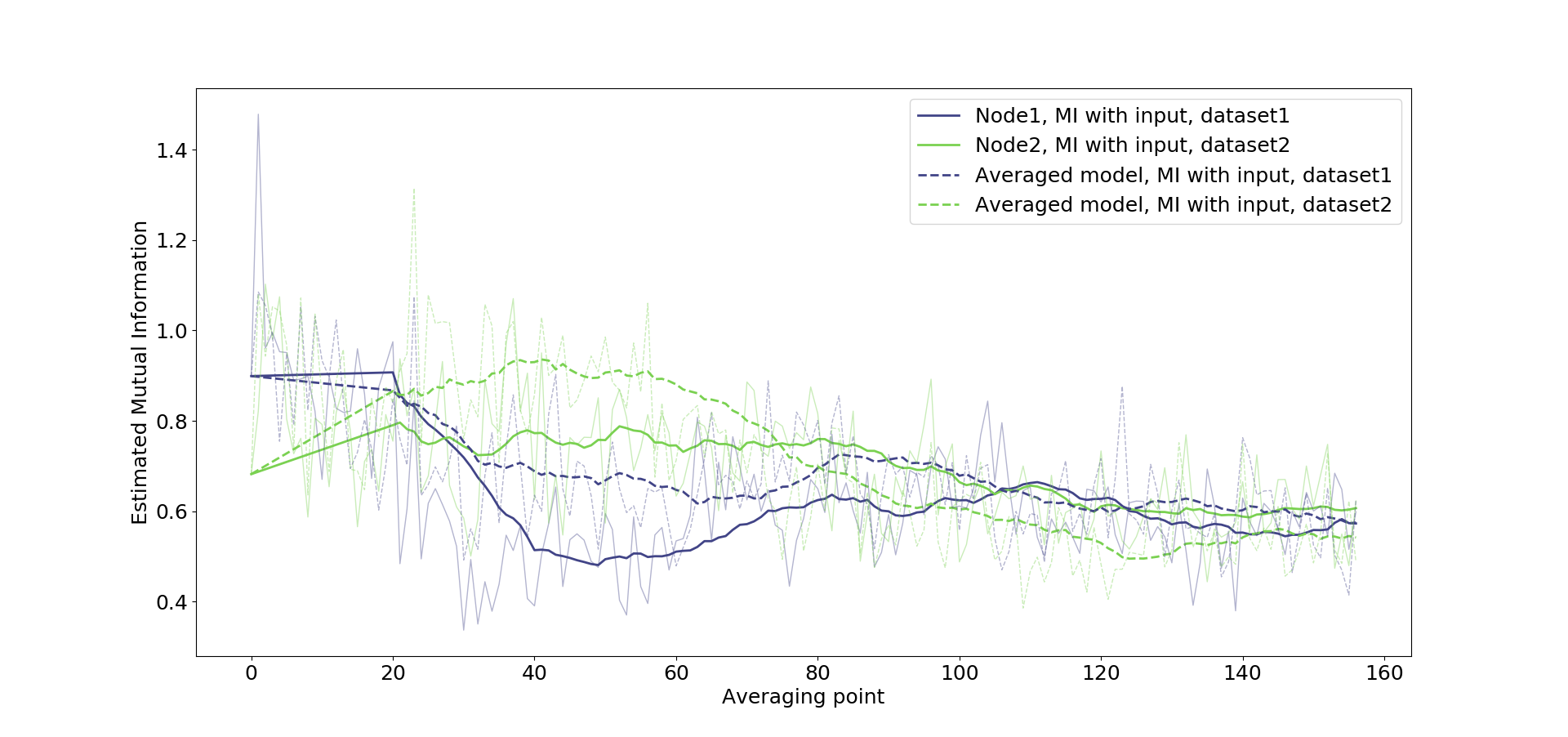}
\endminipage\hfill
\minipage{0.5\textwidth}
  \includegraphics[width=\linewidth]{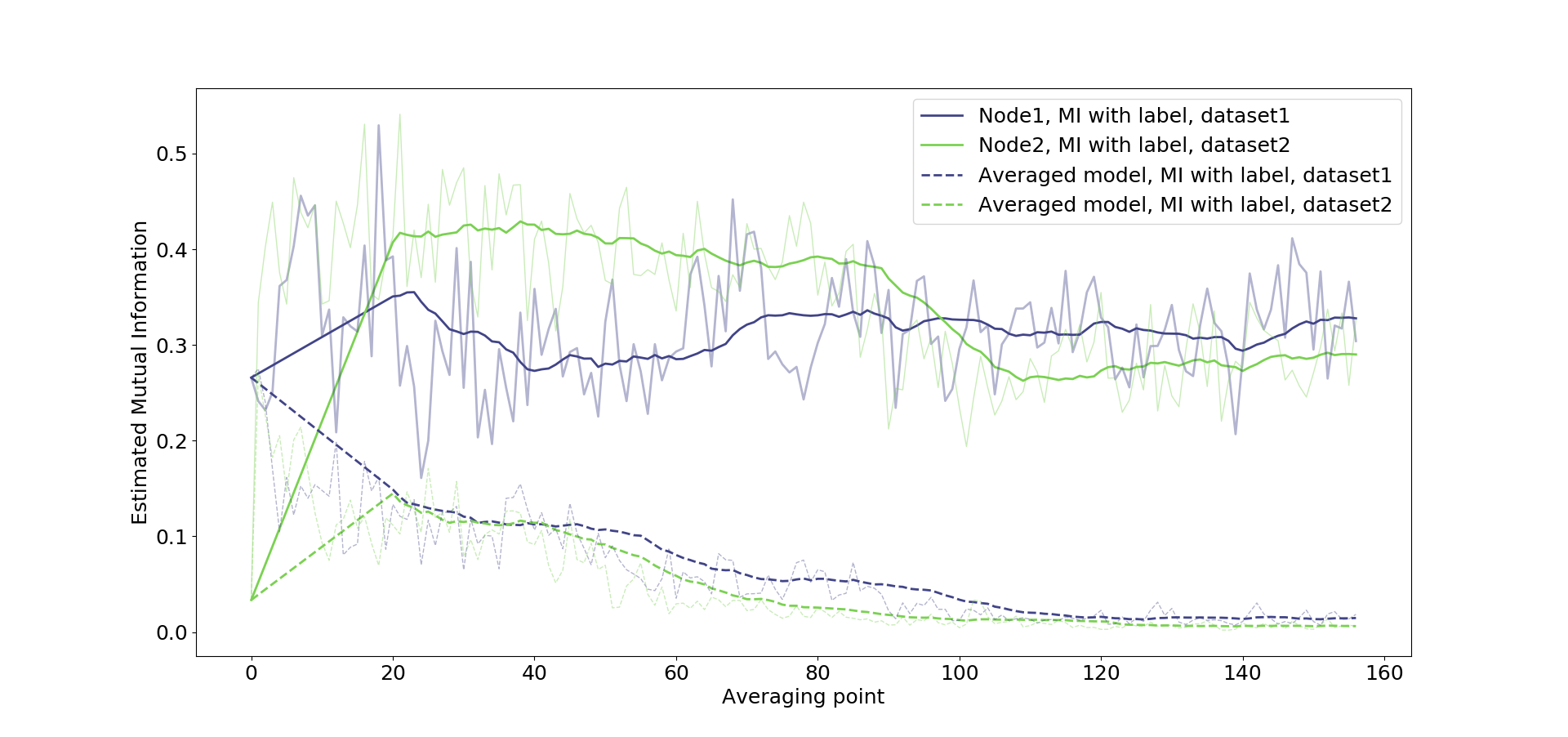}
\endminipage
\caption{Mutual information measured  for the non-iid case without sending back the aggregated model. The estimated amount of information is shown for input $X$ (left) and label $Y$ (right) for each aggregation point.}
\label{fig:noniid_nosync}
\end{figure}

\begin{figure}[ht]
\minipage{0.5\textwidth}
  \includegraphics[width=\linewidth]{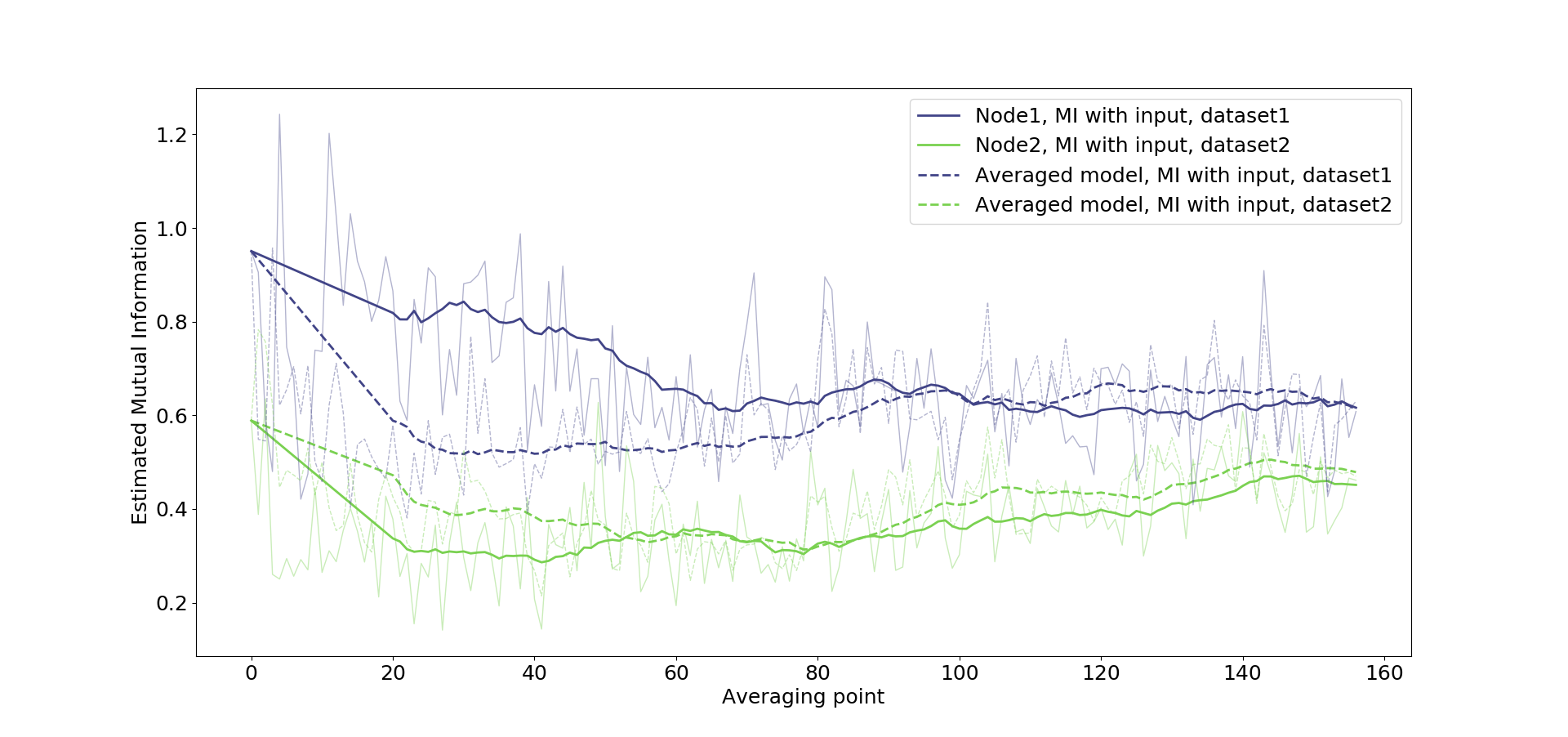}
\endminipage\hfill
\minipage{0.5\textwidth}
  \includegraphics[width=\linewidth]{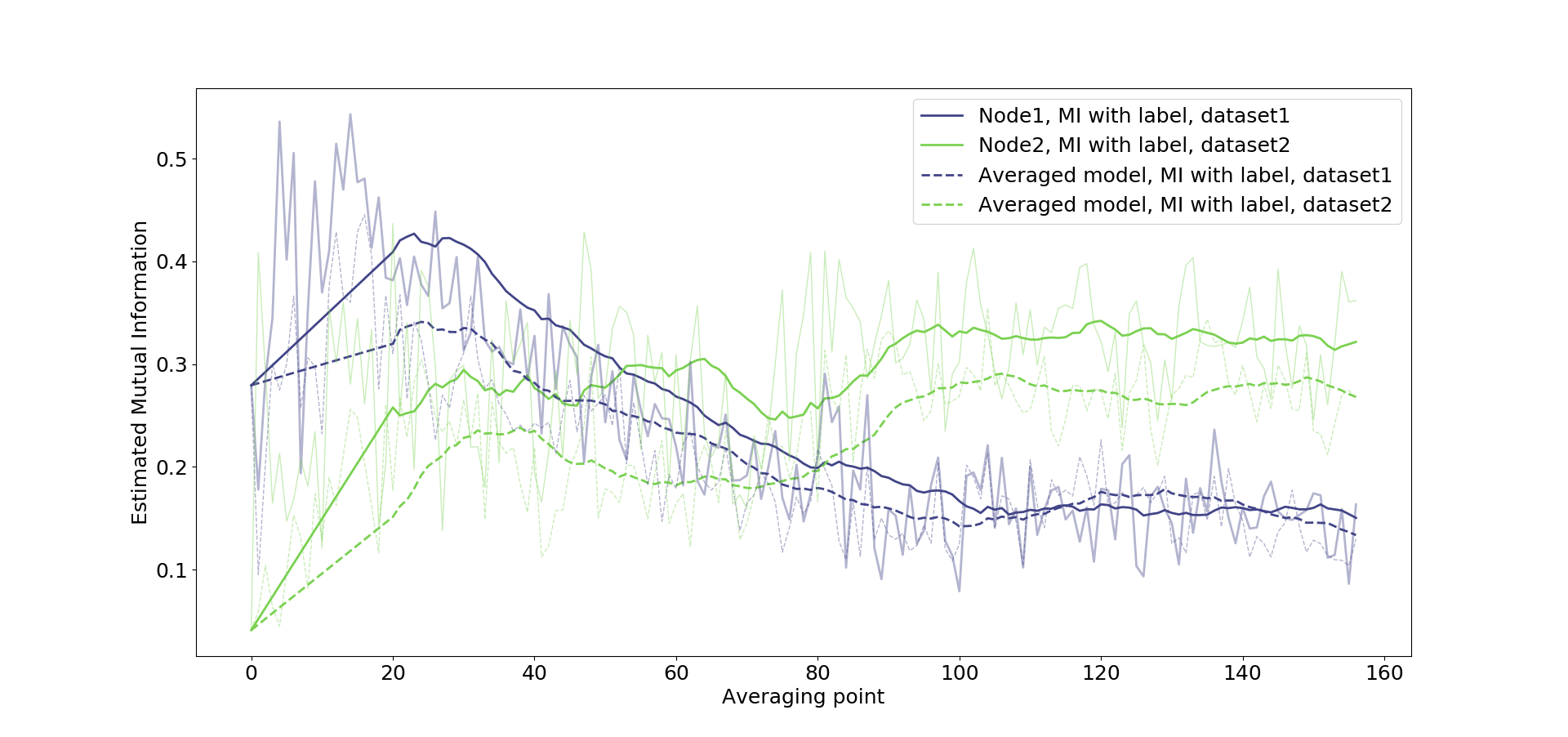}
\endminipage
\caption{Mutual information measured for the non-iid case with aggregation process and continuing learning from the averaged point. The estimated amount of information is shown for input $X$ (left) and label $Y$ (right) for each aggregation point.}
\label{fig:noniid_sync}
\end{figure}

We now repeat the experiments with non-iid local datasets. While with iid datasets the averaged model ceases to be able to combine information about local labels when the local models become more fit, with non-iid datasets this is happening almost instantly. In Figure~\ref{fig:noniid_nosync} on the right the mutual information of the representation and the label estimated for the averaged model decreases from the very beginning of the training process and goes down to $0$. 
%Finally, while local models are much stronger than in the iid case since their task is easier --both achieve more than $88\%$ training accuracy-- the final averaged model has $12\%$ accuracy on the first dataset and $26\%$ on the second. The difference might be explained by the easiness of the second dataset compared to the first one.
%
Figure~\ref{fig:noniid_sync} shows the development of mutual information for non-iid datasets and periodic synchronization. Here we have chosen the period of synchronization $100$ local batches that corresponds to the very first aggregation in Figure~\ref{fig:noniid_nosync}. We do not observe anymore the drop of the mutual information with the labels in the averaged network as in Figure~\ref{fig:noniid_nosync}. Moreover, the mutual information in the averaged model with each of the local datasets stays close to the mutual information estimated for the local nodes. Interestingly enough, it seems that information propagated through averaging from the second node to the first one hinders it from training better. But overall here the final averaged model achieves considerably high accuracy  -- approximately $60\%$ on each of the local datasets. 
%It should also be noted that compared to the first case, training accuracy on the opposite dataset grew from being $0.0\%$ to more than $1\%$, which is especially interesting since each local node has never seen any of the examples of the classes for another node.
%
The analysis of the non-iid case indicates that the very frequent aggregation can be beneficial, but can also deteriorate the local training process.

An inspection of the mutual information with the representation of the aggregated model helps to identify a beneficial synchronization period in terms of training quality and communication costs and also sheds light on the dynamics of the training process with non-iid datasets.

\section{Future Work}
%Our experiments so far have led to some interesting observations about information propagation through the averaging process. In the future, we would like to formalize our application of information theory to the setting of federated learning for deep neural networks with the goal of grounding the experimental results theoretically. In particular, we want to theoretically describe the changes in mutual information during aggregation of local models.

The paper leaves as an open question the theoretical analysis of mutual information development, that can be used to answer the questions of when and how to aggregate local models. The first aspect is closely linked to achieving communication benefits in the sense of reduced communication costs due to an improved timing of aggregation in the federated setup. Further examining how different aggregation methods behave in terms of the development of mutual information can help answering the question of how to optimally aggregate the local models to improve the quality of the resulting global model.

\paragraph{Acknowledgments}
This research has been partly funded by the German Federal Ministry of Education and Research, ML2R - F\"{o}rderkennzeichen 01S18038B.

\newpage
\bibliography{literature}

\end{document}